\documentclass[preprint,review,12pt]{elsarticle}

\usepackage{amssymb}
\usepackage{tikz,graphics,color,fullpage,float,epsf,caption,subcaption}
\usepackage{bm}
\usepackage{amsmath}
\usepackage{icomma}
\usepackage{ulem,xpatch}
\usepackage{hyperref}
\usepackage{appendix}
\usepackage{booktabs}
\usepackage{amsmath}
\usepackage[margin=1in]{geometry}
\usepackage{makecell}

\newcommand{\argmin}{\operatornamewithlimits{argmin}}

\journal{Aerospace Science and Technology}

\begin{document}

\begin{frontmatter}



\title{Goal-Driven Adaptive Sampling Strategies for Machine Learning Models Predicting Fields}


\author[inst1,inst2]{Jigar Parekh}
\author[inst1,inst2]{Philipp Bekemeyer}

\affiliation[inst1]{organization={Cluster of Excellence SE$^2$A – Sustainable and Energy-Efficient Aviation},
            addressline={TU Braunschweig}, 
            city={Braunschweig},
            country={Germany}}

\affiliation[inst2]{organization={Institute of Aerodynamics and Flow Technology},
            addressline={German Aerospace Center (DLR)},
            city={Braunschweig/Göttingen},
            country={Germany}}
\begin{abstract}
    Machine learning models are widely regarded as a way forward to tackle multi-query challenges that arise once expensive black-box simulations such as computational fluid dynamics are investigated. However, ensuring the desired level of accuracy for a certain task at minimal computational cost, e.g. as few black-box samples as possible, remains a challenges. Active learning strategies are used for scalar quantities to overcome this challenges and different so-called infill criteria exists and are commonly employed in several scenarios. Even though needed in various field an extension of active learning strategies towards field predictions is still lacking or limited to very specific scenarios and/or model types. In this paper we propose an active learning strategy for machine learning models that are capable if predicting field which is agnostic to the model architecture itself. For doing so, we combine a well-established Gaussian process model for a scalar reference value and simultaneously aim at reducing the epistemic model error and the difference between scalar and field predictions. Different specific forms of the above-mentioned approach are introduced and compared to each other as well as only scalar-valued based infill. Results are presented for the NASA common research model for an uncertainty propagation task showcasing high level of accuracy at significantly smaller cost compared to an approach without active learning.
\end{abstract}


\begin{graphicalabstract}
\end{graphicalabstract}

\begin{highlights}
\item A model-agnostic active learning strategy for field predictions that extends classical infill criteria beyond scalar outputs.
\item A unified scalar–field surrogate training framework that jointly reduces epistemic uncertainty and enforces consistency between global coefficients and distributed values.
\item Significant accuracy gains at reduced computational cost, demonstrated on the NASA Common Research Model for uncertainty propagation.
\end{highlights}


\end{frontmatter}



\section{Introduction}\label{sec:Introduction}
    High-fidelity aerodynamic simulations (e.g. CFD solutions of the flow around an aircraft) provide detailed flow field information and accurate performance metrics, but they remain computationally expensive especially if multiple parameter combination are to be investigated. Surrogate modeling has become a cornerstone in aerodynamic design to at least partially alleviate this cost. Data-driven models such as Gaussian process regressions, reduced-order models (ROMs), or machine learning and deep learning models, such as graph neural networks (GNNs) are trained to approximate the simulation outputs, trading a high upfront simulation cost for a fast predictive model.  A critical challenge is to ensure that the surrogate is sufficiently accurate with a limited number of training samples (simulation runs). The design and analysis of computer experiments is well established in the literature \cite{Kennedy2001,Santner2003,Forrester2008} providing a solid basis with samples spread of the design space. However, domain specifics like the difference between subsonic and transonic flow are unknown to such design of experiment techniques and, hence, can not be accounted for. Adaptive sampling, or active learning, addresses this by iteratively selecting new sample points in the design or parameter space that are expected to maximally improve the surrogate. For scalar objectives – for example, predicting a single quantity like a drag coefficient or lift-to-drag ratio – there are well-established infill (acquisition) criteria that guide this selection. A classic approach is the Efficient Global Optimization (EGO) algorithm using the expected improvement (EI) criterion \cite{Jones1998,Sacks1989}, which balances exploration and exploitation by favoring points likely to yield a better optimum of the objective. Alternatively, when the goal is accurate surrogate modeling for uncertainty quantification (UQ) rather than purely optimization, one can target reductions in the surrogate’s prediction error across the input distribution. In Gaussian process (GP) surrogates, for instance, methods like Active Learning Cohn (ALC) choose new samples that would maximally reduce the integrated predictive variance of the model \cite{Bect2012}. Such variance-based criteria (related to integrated mean squared error) seek to minimize overall surrogate uncertainty with respect to the input probability density function, ensuring the model is accurate on average for UQ analysis \cite{Gramacy2009,Brochu2010,Shahriari2016}. In summary, for single-output surrogates, techniques like EI and variance reduction provide principled ways to improve surrogate fidelity with minimal samples \cite{Bect2012,Gramacy2009}. These methods have been successfully applied in aerodynamic optimization and analysis, where adaptively sampling the design space yields faster convergence to optimal designs and more reliable surrogate predictions than space-filling designs alone.

    However, many problems in aerodynamic design and analysis demand field predictions rather than just scalar outputs. For example, predicting the pressure or velocity field around an airfoil or across a wing’s surface is crucial for understanding shock locations, flow separation, and structural loads --- information that a single scalar, e.g. the drag coefficient, cannot capture. Building surrogates for such distributed outputs which can be thought of as high-dimensional vectors or functions defined on a spatial domain is far more complex. First, the dimensionality of the output -- potentially thousands of values describing a field -- often necessitates some form of dimensionality reduction. Common approaches include reduced-order modeling techniques such as proper orthogonal decomposition (POD) to represent the field in terms of a few dominant modes or the mapping of a latent variable to the field \cite{Peherstorfer2018, Xiu2010}. Alternatively, one may use deep learning models such as autoencoders that learn a compressed latent representation of the field. Second, even if a surrogate can predict a field, adaptively sampling to improve a field prediction surrogate is not straightforward. In contrast to a scalar objective, a field has many features and possible sources of error, e.g. a surrogate might accurately predict overall trends but miss a local shock entirely. Hence, there is a need for field-based infill criteria that guide the selection of new simulation points by considering the surrogate’s performance on the entire field output. Some studies have begun to explore this; for instance, adaptive sampling strategies for POD-based surrogates have been proposed to iteratively fill the parametric space and improve the surrogate accuracy \cite{Guenot2013}. Furthermore, vector-valued Gaussian processes, which extend standard GP models to directly handle multi-output fields and quantify uncertainty over them, have been developed to drive max-variance based infill strategies \cite{Alvarez2012,Picheny2013}. However, proposed techniques are either limited in scope or tailored towards a certain model type. 

    Extending adaptive sampling to field prediction tasks raises several key challenges.
    \begin{enumerate}[i)]
        \item \textbf{Scalarizing field outputs:} Unlike a single objective, a field output must be summarized or reduced to a one-dimensional criterion for decision-making. Determining an appropriate scalarization (for example, an integrated error norm over the field, a worst-case error, or a weighted sum of errors at critical locations) is non-trivial and can bias the sampling toward certain regions of the input space. Multiobjective optimization research offers many scalarizing functions to combine multiple objectives into one \cite{Chugh2019}, but those methods typically handle a handful of objectives, not the thousands of degrees of freedom in a field. Applying a naive scalarization to a field (e.g. simply using the average error) might overlook localized phenomena, whereas focusing only on the maximum error might overweight outliers. Thus, defining a scalar criterion that truly reflects surrogate field accuracy is an open question.
        \item \textbf{Managing epistemic uncertainty:} Surrogate models for fields must capture uncertainty in high-dimensional outputs. For example, Gaussian process surrogates require multi-output covariance structures to estimate uncertainties and their correlations \cite{Alvarez2012}. In practice, reduced bases with probabilistic models or neural network ensembles are used. However, epistemic uncertainty arises not only from sparse data but also from uncertain model parameters such as correlation lengths or basis truncation. Graham and Cortés \cite{Graham2009} demonstrate that accounting for covariance uncertainty can significantly impact predictive variance, driving adaptive sampling via both model and conditional uncertainties. Ignoring these factors may lead to misinformed infill decisions; hence, rigorous statistical modeling is essential to correctly identify where the surrogate is truly uncertain \cite{Cressie1993,Stein1999}.
        \item \textbf{Consistency between scalars and fields:} Design often requires both global coefficients, e.g. lift and drag, and detailed field information ,e.g. pressure or shear distribution. Training separate surrogates for scalars and fields raises the challenge of allocating new samples optimally. For example, improving drag prediction might not reduce pressure distribution errors, and vice versa. This leads to a multi-output active learning problem where one must balance competing objectives. Focusing solely on one aspect may waste samples or leave the other under-trained, and existing multiobjective infill methods still require reducing field error to a single metric, which does not scale well with multiple outputs \cite{Bect2012}. Achieving an optimal training set that accurately captures both scalar and field outputs is challenging and underexplored in current literature. Traditional multiobjective infill methods can treat, for example, scalar and field error as competing objectives; however, they require reducing field error to a single number, which does not scale well when multiple scalar metrics or field features are of interest.
    \end{enumerate}

Given these challenges, current approaches to surrogate modeling for aerodynamic fields show significant gaps. Bayesian multiobjective optimization techniques can scalarize a few objectives \cite{Chugh2019}, but they do not extend well to high-dimensional field outputs. Meanwhile, spatial sensor placement studies \cite{Graham2009} focus on field estimation under specific covariance assumptions and do not incorporate concurrent optimization objectives or engineering constraints. There is a clear need for a unified, adaptive sampling strategy that is both field-aware and objective-aware, while rigorously accounting for surrogate uncertainty \cite{Sullivan2015} and at the same time being agnostic to the specific machine learning model employed. To the best of our knowledge, no current methodology fully addresses the challenge of efficiently learning both scalar performance metrics and high-dimensional field outputs.
    
In this work, we propose field-based adaptive sampling strategies for surrogate modeling that extend to UQ, optimization, reliability analysis and general data-base generation. We develop novel infill functions that generalize classical criteria, e.g. surrogate error or integrated variance reduction, to field outputs. This criterion effectively balances exploration, i.e. improving the surrogate where it is most uncertain in the flow field, and exploitation, i.e. refining predictions of important global quantities, enabling active learning which may serve various needs. We further present a unified active learning framework that jointly trains surrogates for scalar metrics and field distributions. This infill approach ensures that new samples reduce overall epistemic uncertainty in both scalar and field outputs. It is applicable to any model predicting fields independent of its specific architecture are model structure. The proposed approach enables efficient and robust training of surrogates that can predict full flow fields with quantified uncertainty, dramatically reducing the number of expensive simulations needed. This leads to surrogates that remain accurate across the input distribution and reliable for downstream tasks such as global sensitivity analysis, probabilistic design optimization, and risk assessment in aerodynamic applications. 


The paper is structured as follows: Section \ref{sec:Methodology} introduces the goal-driven adaptive sampling strategies for surrogate models predicting fields. Section \ref{sec:Setup} details the setup for NASA CRM simulations and surrogate modeling; including problem formulation, flight conditions, transition modeling, environmental and operational parameters, and the numerical model for computing aerodynamic quantities of interest (QoIs). Section \ref{sec:Results} presents the results from different infill approaches and the application of best-of-all model for the UQ analysis. Finally, conclusions are offered in section \ref{sec:Conclusion}.

\section{Methodology}\label{sec:Methodology}
    
    In this study, surrogate models are developed for both scalar outputs and spatially distributed field outputs. The adaptive sampling proposed accommodates both the surrogates directly estimate quantity of interest (QoI).
    The following sections detail the surrogate modeling methods and adaptive infill criteria employed to ensure high predictive accuracy in both scalar and field contexts. Even though results are later on presented for an aerodynamic use-case the methodology is applicable to any discipline or setting as long as the field prediction can be integrated into an engineering-feasible scalar quantity that than be modelled using a Gaussian Process. 
    \subsection{Surrogate Modeling without Infill}
        We first construct baseline surrogate models using an initial design of experiments (DoE) before any adaptive refinement. Three surrogate modeling techniques are employed, reflecting the diversity of scalar vs. field outputs.

        \subsubsection{Gaussian Process Regression for Scalars}
            Gaussian Process (GP) models are employed to surrogate scalar quantities of interest (QoIs), due to their flexibility and intrinsic uncertainty estimation capability~\cite{bishop2006pattern}. The true response \(y(\mathbf{x})\) is assumed to be a realization of a random function \(Y(\mathbf{x})\), composed of a deterministic trend and a zero-mean GP noise term. Formally, this can be expressed as:
            \begin{equation}
            Y(\mathbf{x}) = g(\mathbf{x})\boldsymbol{\beta} + \epsilon(\mathbf{x}),
            \end{equation}
            where \(g(\mathbf{x})\boldsymbol{\beta}\) represents a known regression model (trend) and \(\epsilon(\mathbf{x})\) is a GP with covariance given by
            \begin{equation}
            \mathrm{Cov}[\epsilon(\mathbf{x}),\epsilon(\mathbf{x}')] = \sigma^2 R_{\boldsymbol{\theta}}(\mathbf{x},\mathbf{x}'),
            \end{equation}
            using stationary covariance kernels, e.g., Gaussian-exponential family, with hyperparameters \(\boldsymbol{\theta}\) estimated via maximum likelihood~\cite{Bertram2013}. 
                    
            The GP predictor provides interpolation of training data and posterior variance estimates, yielding the best linear unbiased estimator:
            \begin{equation}
            \hat{y}(\mathbf{x}) = g(\mathbf{x})\boldsymbol{\beta} + \mathbf{r}(\mathbf{x})^T R^{-1}(Y - F\boldsymbol{\beta}),
            \end{equation}
            where \(R\) is the correlation matrix of training points, \(\mathbf{r}(\mathbf{x})\) is the correlation vector between \(\mathbf{x}\) and training points, \(Y\) are observed outputs, and \(F\) is the design matrix. The GP further estimates the mean-squared error (MSE) at any input \(\mathbf{x}\) \cite{Nymand2002}:
            \begin{equation}
            s^2(\mathbf{x}) = \sigma^2\left(1 - \mathbf{r}(\mathbf{x})^T R^{-1}\mathbf{r}(\mathbf{x}) + \mathbf{u}^T(F^T R^{-1}F)^{-1}\mathbf{u}\right),
            \end{equation}
            with \(\mathbf{u} = F^T R^{-1}\mathbf{r}(\mathbf{x}) - g(\mathbf{x})\). This \(s(\mathbf{x})\) acts as a local surrogate error indicator, making GPs particularly suited for uncertainty quantification.
    
        \subsubsection{Proper Orthogonal Decomposition with Interpolation for Fields}
            For high-dimensional field outputs, Proper Orthogonal Decomposition with Interpolation (PODI) can be utilized~\cite{Guo2019}. POD generates an efficient representation of output fields by expressing the field as a linear combination of orthonormal modes. Given an initial set of \(N_0\) field snapshots \(\{u(\mathbf{x}_k)\}_{k=1}^{N_0}\), each of dimension \(p\gg 1\), singular value decomposition (SVD) yields a reduced basis \(\{\phi_i\}_{i=1}^M\) capturing the most significant variations \cite{Eckart1936}. Any field can be approximated as:
            \begin{equation}
            u(\mathbf{x}) \approx u_0 + \sum_{i=1}^{M} \alpha_i(\mathbf{x})\phi_i,
            \end{equation}
            where \(u_0\) is the mean field and \(\alpha_i(\mathbf{x})\) are the mode amplitudes dependent on inputs \(\mathbf{x}\). The problem reduces to modeling the mapping from inputs to mode amplitudes, typically via multi-dimensional regression techniques like GPs or Radial Basis Functions (RBFs)~\cite{Guo2019}. This PODI approach significantly reduces output dimensionality, providing efficient and accurate field predictions, as demonstrated by recent studies.
    
        \subsubsection{Graph Neural Networks for Fields}
            Graph Neural Networks (GNNs) offer a powerful data-driven alternative for modeling complex spatial fields, particularly suited for outputs defined over unstructured meshes common in aerodynamic and fluid dynamics problems \cite{Pfaff2021, Hines2023}. GNNs interpret mesh points as graph nodes and mesh connections as graph edges, utilizing iterative message-passing algorithms to learn intricate spatial interactions directly from data~\cite{Gonzalez2020,Gilmer2017}.
            
            The message-passing mechanism in GNNs generally follows an iterative update process. For each node $i$, the latent feature representation $\mathbf{h}_i^{(l)}$ at the $l$-th iteration is updated according to:
            \begin{equation}
                \mathbf{m}_i^{(l+1)} = \sum_{j \in \mathcal{N}(i)} \phi\left(\mathbf{h}_i^{(l)}, \mathbf{h}_j^{(l)}, \mathbf{e}_{ij}\right),
            \end{equation}
            \begin{equation}
                \mathbf{h}_i^{(l+1)} = \psi\left(\mathbf{h}_i^{(l)}, \mathbf{m}_i^{(l+1)}\right),
            \end{equation}
            where $\mathcal{N}(i)$ denotes the set of neighboring nodes to node $i$, $\mathbf{e}_{ij}$ represents edge features, and $\phi$ and $\psi$ are neural network functions typically realized as multi-layer perceptrons (MLPs). For further details on the employed GNN methodology the interested reader is referred to \cite{Hines2023}.
    
            The main advantages of GNN-based approaches include inherent flexibility in mesh topology representation, the ability to directly incorporate geometric and topological information into learning, and enhanced accuracy in capturing spatial correlations compared to traditional surrogate modeling techniques. These capabilities make GNNs a compelling choice for surrogate modeling tasks in aerodynamics and fluid mechanics.
        

    \subsection{Adaptive Sampling (Infill) Approaches}
        The adaptive sampling (infill) step is solely governed by the acquisition function which decides how the next sample is determined. 
        For applications with scalar output the next sample point $\bm{\xi}^*$ is selected by maximizing the criteria value $I(\bm{\xi})$:
        
        \begin{equation} 
            \bm{\xi}^* = \argmin_{\bm{\xi}} \{ -\, I(\bm{\xi}) \, \}
        \end{equation}

        A typical and simple choice for the criteria is the surrogate error, i.e. $I(\bm{\xi})=\hat{s}(\bm{\xi})$. Such error-based adaptive sampling has been shown to greatly improve efficiency over unguided MC sampling \cite{Sabater2020,Parekh2024}. The optimization of the acquisition function is performed with a global optimizer (here, differential evolution) which is effective in navigating complex, multimodal objective landscapes.

        It should be noted that, for applications with non-uniformly distributed inputs the next infill point is typically selected by maximizing the product of the joint input probability density function of the inputs $\text{PDF}_{\xi}$ and the respective criteria value, i.e. $\bm{\xi}^* = \argmin_{\bm{\xi}} \{ -\, \text{PDF}(\bm{\xi})I(\bm{\xi}) \, \}$. This criterion prioritizes sampling in high-probability regions while simultaneously targeting regions with high criteria value. 
    
        \subsubsection{Adaptive Sampling for Scalars}
            The surrogate-error criterion works well for single-output (scalar) GP surrogates as these have an inherent estimation of model uncertainty i.e. $\hat{s}(\bm{\xi}) = \sigma_{GP}(\bm{\xi})$. We will refer to this approach as \textit{GP Infill:SE}. Note that this approach still needs a surrogate model for field predictions. We employ a PODI or GNN using together the DoE and the infill samples of the GP Infill:SE approach to construct a field surrogate enabling inference at new inputs.
            
            While using a scalar GP’s error is convenient, it may not fully capture where the field surrogate itself has high error. For instance, the pressure distribution might be poorly predicted in some region of input parameter space that does not greatly affect the integrated scalars $C_D$ and/or $C_L$. Infill purely based on $\sigma_{GP}$ may not explore such regions.

        \subsubsection{Adaptive Sampling for Fields}
            For deep learning field surrogates, we must extend the concept of SE-based infill. A straightforward approach is to derive a scalar metric from the field and then apply a similar criterion. However, unlike GPs,  GNNs lack inherent uncertainty estimates. To circumvent this issue, Monte Carlo dropout or ensemble methods can be used to approximate predictive uncertainty~\cite{Gal2016,Breiman1996}. In this study we employ the former, the details of which are deferred to \ref{app:GNNEpUnc}. Using the approximated model variance as the SE indicator i.e. $\hat{s}(\bm{\xi}) = \sigma_{GNN}$, we can apply the SE-based infill just like we did with GPs effectively reducing the field infill problem to the scalar case. We refer to this approach as \textit{GNN Infill:SE}.
    
            Just like GP, the GNN surrogate is retrained after every infill iteration. Therefore, the GNN surrogate keeps changing in terms of its prediction and its variance at each training sample after every infill iteration. One potential problem with this is that it may not guarantee a significant reduction in the model's epistemic uncertainty even after large number of infill steps.

        \subsubsection{Coupled Adaptive Sampling}
            \begin{figure}[t]
                \centering
                \includegraphics[trim={0 0 0 0}, scale=0.8, clip]{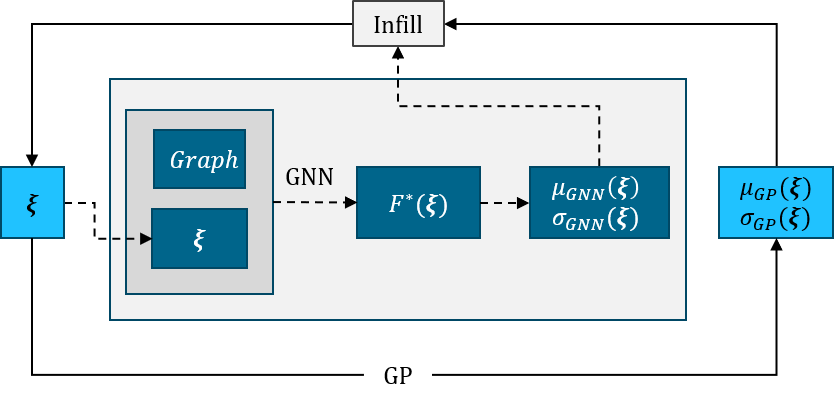}
                \caption{Coupled goal-driven scalar-field adaptive sampling framework. A Gaussian process (GP) surrogate maps the inputs $\bm{\xi}$ to scalar coefficients resulting in a mean and variance approximation denoted by $\mu_{GP}$ and $\sigma_{GP}$, while a field model receives the same inputs together with other additional information such as the surface-mesh graph and predicts field quantities $F^*$ from which the scalars are integrated out with thereby estimated scalar mean and variance denoted by $\mu_{GNN}$ and $\sigma_{GNN}$. The misfit and epistemic uncertainties of both surrogates feed a coupled infill criterion that selects the next design point. The new snapshot closes the loop and updates both surrogates.}
                \label{fig:GP-GNN}
            \end{figure}

            To simultaneously improve the accuracy of scalar and field surrogates, we introduce two acquisition strategies that couple the scalar surrogate and field surrogate predictions. 
    
            The first criterion augments the GP’s uncertainty with the misfit between the GP and field surrogate predictions. At a given $\mathbf{x}$, let $\mu_{GP}(\mathbf{x})$ be the GP-predicted QoI ( and $\mu_{GNN}(\mathbf{x})$ be the corresponding QoI computed from the GNN-predicted field $F^*(\bm{\xi})$. We define the misfit $\Delta(\bm{\xi}) = |\mu_{GP}(\bm{\xi}) - \mu_{GNN}(\bm{\xi})|$ as a measure of disagreement between the two surrogates. The acquisition function is then given by:
    
            \begin{equation} 
                \bm{\xi}^* = \argmin_{\bm{\xi}} \{- \, [\sigma_{GP}(\bm{\xi}) + \lambda \Delta(\bm{\xi})] \, \}
            \end{equation}
    
            Here $\lambda$ is a weighting factor (we take $\lambda=1$ for simplicity). We refer to this approach as surrogate error with misfit; \textit{GP-GNN Infill:SEwMisfit}. This method essentially prioritizes points that either have a large GP predictive uncertainty or a large discrepancy between the GP and field surrogate. In applications with non-uniform inputs, as discussed earlier, the criteria value is weighted by the input PDF to remain focused on probable regions. This can accelerate learning for the field surrogate - by adding points where the field prediction disagrees with the well-trained GP, we specifically target improvements in the field model. SEwMisfit is in general related to model discrepancy sampling \cite{Gardner2021} or multi-fidelity infill \cite{Garbo2024}, where one model guides sampling for another.

            The second coupled strategy is more statistically rigorous. We consider the predictive probability distributions of the GP and the field surrogate for a chosen QoI at an input sample. The GP yields a normal distribution $P_{\bm{\xi}} = \mathcal{N}(\mu_{GP}(\bm{\xi}),\sigma_{GP}^2(\bm{\xi}))$. For the field surrogate, as mentioned earlier, we can obtain an approximate distribution $Q_{\bm{\xi}} $ e.g. via Monte Carlo dropout ensembles, yielding mean $\mu_{GNN}({\bm{\xi}} )$ and variance $\sigma_{GNN}^2({\bm{\xi}})$. We then use the Jensen–Shannon divergence \cite{Lin1991} between these two distributions as the acquisition metric. The Jensen–Shannon divergence is a symmetrized measure of difference between $P$ and $Q$, defined by:
    
            \begin{equation} 
                JSD(P||Q) = \frac{1}{2}KL(P||M) + \frac{1}{2}KL(Q||M)
            \end{equation}        
            
            where $M=\frac{1}{2}(P+Q)$ and $KL$ denotes Kullback–Leibler divergence \cite{Kullback1951}. We evaluate $JSD(P||Q)$ between the two model predictive distributions at each $\bm{\xi}$. The acquisition function is then given by:
    
            \begin{equation} 
                \bm{\xi}^* = \argmin_{\bm{\xi}} \{- \, \text{PDF}_{\xi}(\bm{\xi}) \, JSD(P_{\bm{\xi}}||Q_{\bm{\xi}}) \}
            \end{equation}
    
            This criterion targets points where the entire predictive distributions disagree maximally. We refer to this approach simply as \textit{GP-GNN Infill:JSD}. Unlike SEwMisfit which only uses the mean discrepancy (and GP variance), JSD accounts for cases where, for example, the GNN is very uncertain (broad $Q_{\bm{\xi}}$) while the GP is confident (narrow $P_{\bm{\xi}}$), or vice versa. By sampling such locations, we expect to reduce the epistemic uncertainty of both models. In effect, JSD infill tries to make the GP and field surrogate agree, by adding data until their predictive distributions align in the input space. This strategy has parallels to recent active learning methods in reliability engineering that use JSD to update Kriging models \cite{Chen2021}, demonstrating improved efficiency and convergence in failure probability estimation. However, even though mathematically rigerous with respect to matching distributions it is affected from the same drawbacks with respect to Monte-Carlo dropout discussed before.
    
            Both SEwMisfit and JSD reduce to standard SE when the field surrogate aligns perfectly with the GP. When field surrogate lags in accuracy, they prioritize reducing this discrepancy, while still weighting samples by the input PDF. This results in infill samples that simultaneously improve scalar and field predictions.

            \begin{figure}[t]
                \centering
                \includegraphics[trim={0 0 0 0}, scale=0.6, clip]{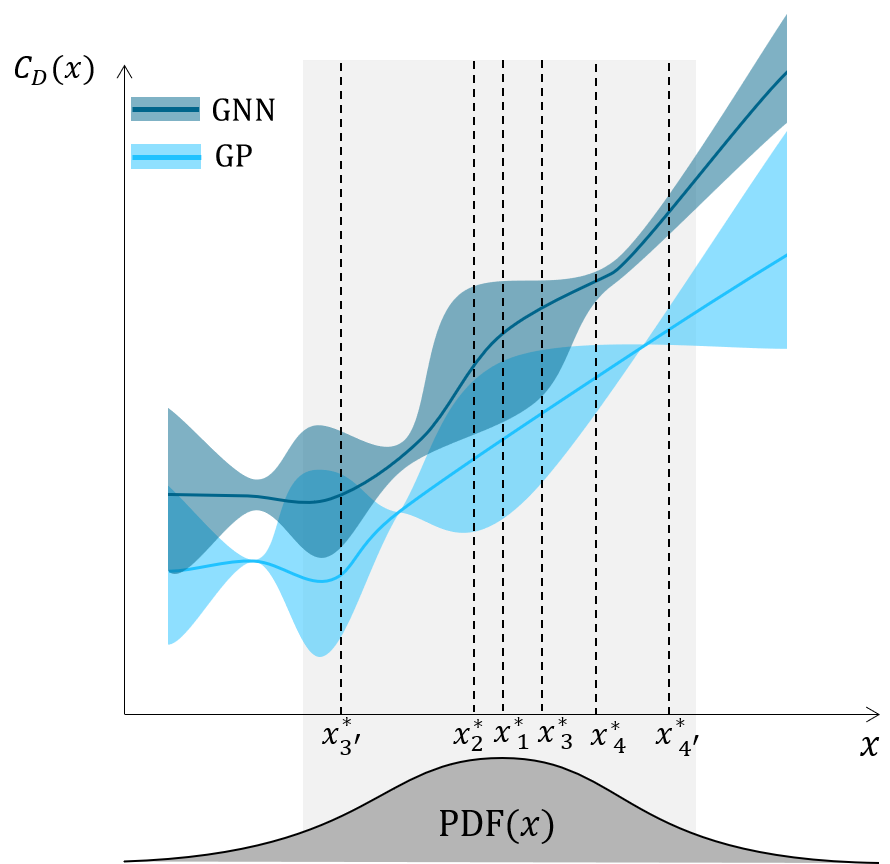}
                \caption{Conceptual comparison of the four adaptive-sampling criteria used for the coupled GP–GNN surrogate system. The solid lines represent the GP (light-blue) and GNN (dark-blue) predictions of the drag coefficient $C_D(x)$; the shaded bands indicate their epistemic uncertainty. The gray bell curve is the joint input probability density PDF$(x)$. Vertical dashed lines mark candidate infill locations $x_i^*$ proposed by each strategy: $x_1^*$ - Surrogate Error (SE) criterion applied to the GP; $x_2^*$ - SE applied to the GNN; $x_3^*$ - SE with GP–GNN misfit (SEwMisfit); $x_4^*$ - Jensen–Shannon divergence (JSD) coupling the two surrogates. The possibility of almost equally probable next infill sample is represented as $x_{i^{'}}^*$.}
                \label{fig:Infill}
            \end{figure}
            
            \autoref{fig:Infill} depicts all adaptive sampling approaches discussed. In the schematic example, the location of the points $x_1^*$ and $x_2^*$ is with respect the Infill:SE criterion for GP and GNN, respectively - sampling where the product of model uncertainty and input distribution is highest. Points $x_3^*$ would typically be chosen based on SEwMisfit criterion. However, point $x_{3^`}^*$ may also chosen based on this criterion - indicating that a sample maybe chosen even in the lower probability region if the sum of the GP uncertainty and the GP-GNN misfit is significant. Similarly, using JSD criterion, sample points $x_{4}^*$ and $x_{4^`}^*$ may be chosen in support of large mismatch between the two surrogates - even in the lower probability regions of the input distribution.




\section{Numerical Setup}\label{sec:Setup}
  
    The aerodynamic configuration considered is the NASA Common Research Model with a Natural Laminar Flow wing (CRM-NLF). This wing was originally designed by Lynde \textit{et al.} \cite{Lynde2017} using a Crossflow Attenuated Natural Laminar Flow (CATNLF) method to delay both Tollmien–Schlichting (TS) and crossflow (CF) instabilities. A semi-span wind tunnel model of the CRM-NLF was later tested at NASA’s National Transonic Facility, confirming the design’s ability to achieve extensive laminar flow at transonic cruise conditions \cite{Lynde2019}. For the present study, we utilize the same geometry and focus on similar flow conditions.

    \subsection{Numerical Model}\label{sec:NumMod}
\begin{figure}[t]
            \centering
            \includegraphics[trim={0 0 0 0}, scale=0.30, clip]{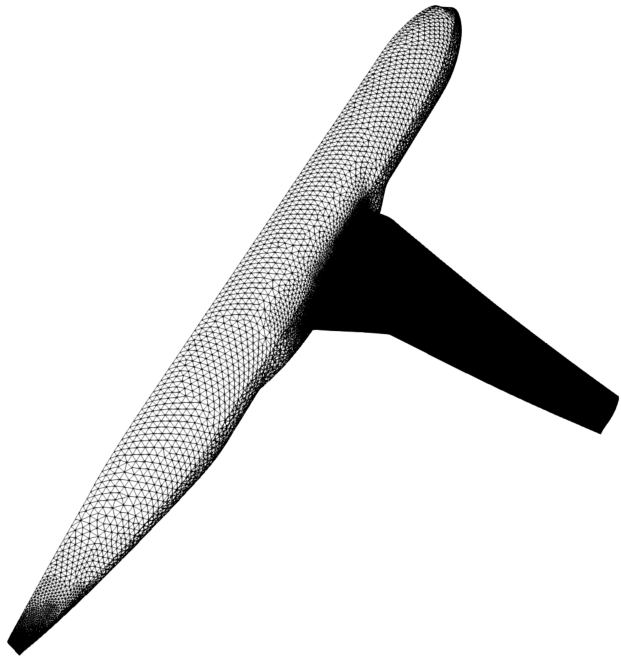}
            \caption{Computational mesh for the NASA CRM.}
            \label{fig:mesh}
        \end{figure}

        The flow around the wing and fuselage NASA CRM configuration is simulated using the TAU CFD solver from the German Aerospace Center (DLR)~\cite{Gerhold2005}. The aerodynamic properties of interest are determined by solving the Reynolds-averaged Navier-Stokes (RANS) equations along with the $k-\omega$ SST turbulence model \cite{Menter2003}. The solver's configuration includes a 4w multigrid cycle, a backward Euler solver for pseudo-time integration and a central flux discretization scheme. The employed unstructured mesh comprises of about 50 million points and about 0.3 million surface points, as shown in Figure \ref{fig:mesh}. Laminarity (transition prediction) is considered for both the upper and the lower surface of the wing plus fuselage configuration.
    
        To predict the transition front over the wing, a recently developed DLR $\gamma-$ model \cite{Francois2023} is used. The class of transition transport models has recently gained significant attention, in particular for application in laminar airfoil design as well as in transition front prediction over a wing \cite{Krumbein2022}. The DLR $\gamma$ model, an enhancement of the $\gamma$-$Re_{\Theta_t}$ model, for transport aircraft applications is a member of this class and used as transition prediction method in this work. It includes the "Simple-AHD" criterion \cite{Perraud2014}, i.e. an adapted  criterion accounting for pressure gradients and considering the effect of compressibility. Furthermore, it includes a crossflow extension \cite{Francois2022b}. It has been effective for laminar wings at high Reynolds numbers  (around $10^7$) and is continuously validated and extended \cite{Helm2023}.

    \subsection{Environmental and Operational Parameters}\label{sec:UncChar}
        \begin{figure}[t]
            \centering
            \begin{subfigure}[t]{0.495\textwidth}
                \centering
                \includegraphics[trim={0 0 0 0}, clip, width=\linewidth]{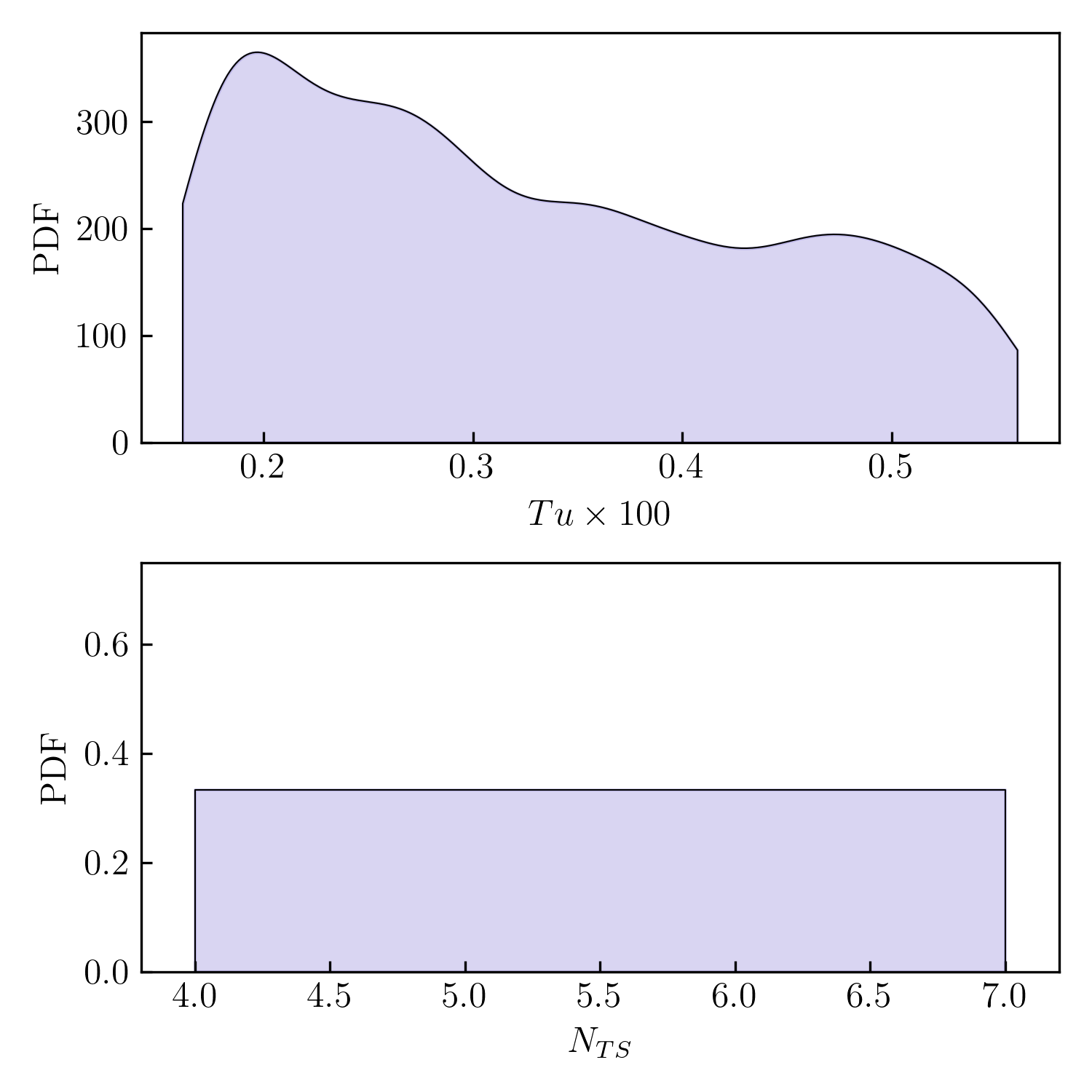}
                \caption{Distribution of $N_{TS}$ and $Tu$.}
                \label{fig:NTS_Tu}
            \end{subfigure}
            \hfill
            \begin{subfigure}[t]{0.495\textwidth}
                \centering
                \includegraphics[trim={0 0 0 0}, clip, width=\linewidth]{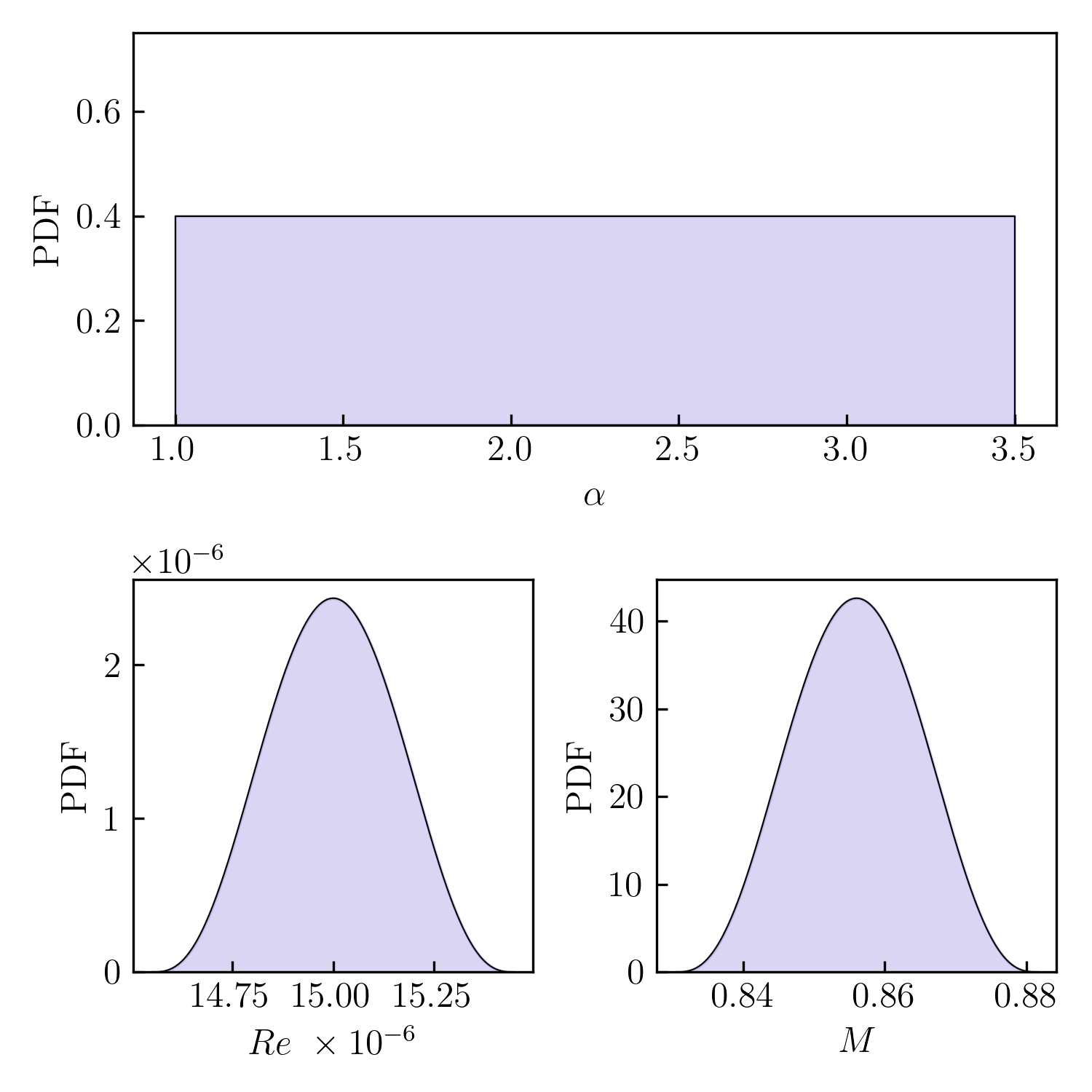}
                \caption{Distribution of $\alpha$, $Re$, and $M$.}
                \label{fig:alphaReM}
            \end{subfigure}
            \caption{(Left) $N_{TS}$ and $Tu$ uncertainty distributions. (Right) Uncertainty distributions for angle of attack, Reynolds number, and Mach number.}
            \label{fig:AllUncertainties}
        \end{figure}

        The goal-based adaptive sampling strategies discussed in \autoref{sec:Methodology} is equally applicable for uniformly or non-uniformly distributed inputs. We consider training the surrogate models based on joint input distribution, as done in previous studies \cite{Sabater2020,Parekh2024}, enabling exploitation of regions with high-probability in the parameter space.  

        To understand the effect of environmental and operational variations on the transition front and pressure distribution over the wing, relevant factors and their effects must be studied. The critical N-factors ($N_{TS}$, $N_{CF}$) used in the $e^N$ transition model \cite{Ingen1956} captures flow quality, significantly influenced by freestream disturbances (e.g., clouds, turbulence intensity), surface imperfections, and acoustic disturbances \cite{Schrauf2004, Streit2015, Kruse2018}. Environmental uncertainties in these N-factors are modeled as uniformly distributed variables, e.g., $N_{TS} \sim \mathcal{U}[5,14], N_{CF}\sim\mathcal{U}[4,11]$ \cite{Sabater2022}. For the DLR $\gamma$ model, uncertainties are represented through variations in freestream turbulence intensity $Tu$, related to $N_{TS}$ by Mack's formula \cite{Mack1977}:
        \begin{equation}\label{eq:NTS_Tu}
            N_{TS} = -8.43 - 2.4\,\ln(Tu).
        \end{equation}
        Sampling $N_{TS}$ from the specified uniform distribution yields the corresponding distribution for $Tu$. The influence of surface roughness (impacting $N_{CF}$) is not included here due to implementation limitations. Distributions for the N-factors and turbulent intensity are shown in Figure~\ref{fig:AllUncertainties} (left). Sampling $N_{TS}$ between 4 and 7 captures the range of critical $N$-factors determined experimentally in the NTF (typical values $5\!-\!6$) versus calmer flight environments ($N_{TS}\approx10$)~ \cite{Lynde2017}. Mapping to $Tu$ via Mack’s formula preserves the physical coupling between modal amplification and disturbance level without committing to a particular analytic PDF.
       
        In many aerodynamic studies the operational conditions such as freestream Mach number, Reynolds number and aircraft angle of attack are prescribed as deterministic inputs.  For short-haul transports, however, continuous throttle adjustments, weather-related altitude excursions and mild gust encounters lead to non-negligible variability in these quantities even during nominal cruise, which in turn alters boundary-layer stability and overall performance \cite{Radespiel2014}. Therefore, in the present study the three quantities are therefore modeled as random variables with carefully chosen probability laws. The operational uncertainty space is described by three independent random variables as shown in Figure ~\ref{fig:AllUncertainties} (right).        
        A uniform distribution of angle of attack $\alpha$ over $[1.0^{\circ},3.5^{\circ}]$ reproduces the small trim adjustments that were necessary in the NTF to match target sectional lift and to compensate for support-system offsets ~\cite{Lynde2019a}. Transition fronts on the CRM-NLF wing shifted by several percent chord within this narrow $\alpha$ band, making the interval critical. 
        To guarantee repeatable flow quality and TSP imaging, the wind tunnel can be assumed to be operated at a Reynolds number which is normally distributed with $\mu=15\times10^{6}$ and a Coefficient of Variation (CoV) of $1\%$ reflecting those tight operational constraints while still allowing viscous-scale variations that affect crossflow growth. 
        Similarly, Mach number is modeled with $\mu=0.856$ and a $1\%$ CoV, capturing small variations which may significantly modify shock strength and therefore the pressure-gradients. 
        These distributions are deliberately conservative (uniform or very narrow distributions) yet encompass the reasonable spread around nominal conditions, recorded in the CRM-NLF data set. They therefore provide a physically consistent and statistically unbiased basis e.g. in a surrogate-based uncertainty propagation.
    
    \subsection{Surrogate Modeling and Adaptive Sampling Setup}\label{sec:SurrModwInfill}
       All surrogate models utilize the same set of four environmental and operational conditions $\bm{\xi}$ as the inputs, i.e. freestream turbulence intensity (Tu), angle of attack ($\alpha$), Reynolds number (Re), and Mach number (M) as inputs. A Gaussian process (GP) surrogate maps the inputs to scalar coefficients $(C_L, C_D)$, while a PODI or GNN receives the inputs and predicts field quantities $(C_p, C_f, C_{f_{x}}, C_{f_{y}}, C_{f_{z}})$ from which the lift and drag coefficients are integrated. In addition the GNN model also has access to the surface-mesh graph.

       The initial design of experiments (DoE) is constructed using Sobol quasi-random sequences ~\cite{Sabater2020} based on prescribed input distributions, ensuring an even coverage of highly probable samples. We consider two configurations: one with a fixed DoE (no infill) and one with adaptive sampling i.e. DoE with infill. In the no-infill case, the surrogate models are trained using the DoE dataset and validated on a separate holdout set, with a further independent test set for final assessment. In the infill case, the models start from the same DoE dataset but are iteratively refined by adding new samples (infill points) selected via an active learning criterion. After each sequential infill iteration, the training set grows by one sample (with the validation and test sets remaining the same), allowing direct comparison. The infill strategy follows the flowchart illustrated in \autoref{fig:GP-GNN}. Surrogate models are trained on the current dataset, their predictive uncertainties (or error indicators) are evaluated across the input space, and the next sample(s) are chosen where the model confidence is lowest or error highest, then the high-fidelity simulation is run at those points and the results appended to the training dataset. This loop is repeated until a budget of infill samples is exhausted. In cases where both a field surrogate and a scalar surrogate are used together (e.g. a GNN for flow field and a GP for an integrated quantity), a coupled infill strategy is employed such that the selection of new points considers uncertainties of either GP or both models so that the added sample improves the accuracy of both field and global scalar prediction. All DoE and Surrogate Models are constructed and trained using the DLR Surrogate Modeling for Aero Data Toolbox (SMARTy) \cite{Bekemeyer2022}



       For the scalar response the GP uses a squared-exponential kernel with a regularization of $1e-12$ to ensure numerical stability. Kernel length-scales and signal variance are determined by maximizing the log-marginal likelihood. The field surrogate GNN uses a ResGatedGraphConv convolution with node feature vector comprising of the normalized spatial co-ordinates appended by the four input variables. Four message-passing layers with 64 nodes each are followed by a linear head that output surface pressure and friction coefficient fields. Inputs are min–max scaled, targets standardized, activations use ReLU with $5\%$ dropout, and training employs Adam (learning rate $lr=1e-3$ with an exponential decay to $5e-4$). The initial DoE surrogate is trained for 3000 epochs. Each infill retraining typically converges in $100-300$ epochs, with an early-stopping patience of 20 epochs based on validation RMSE.
        
\section{Adaptive Sampling Results}\label{sec:Results}
    In the upcoming subsections, we will discuss the performance of both no-infill and adaptive sampling based modeling approaches described in the previous sections, followed by the application of the best model for surrogate-based uncertainty propagation. 

    \subsection{Surrogate Models without Infill}
    \begin{figure}[t]
        \centering
        \includegraphics[trim={0 0 0 0}, scale=0.55, clip]{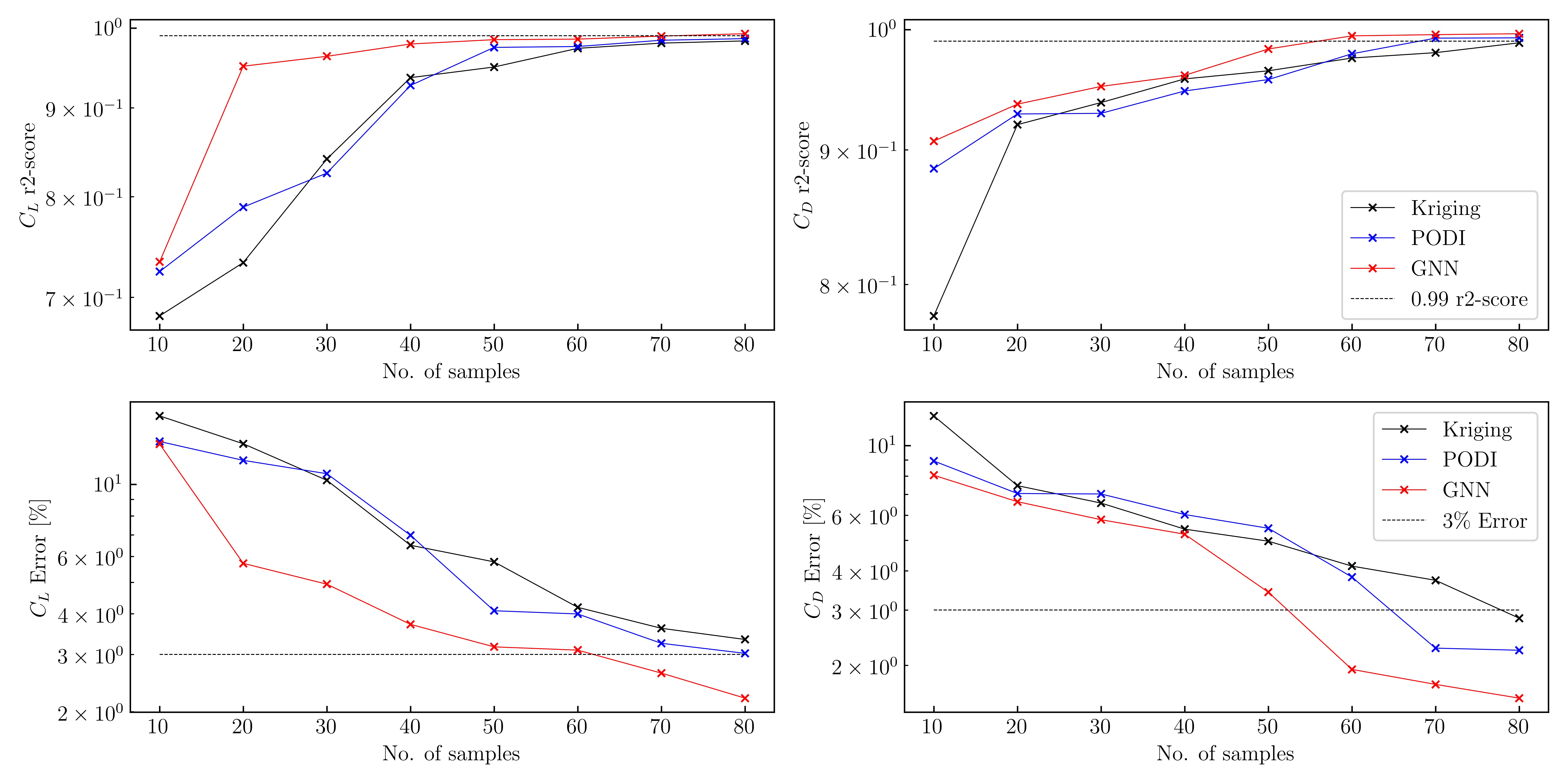}
        \caption{Convergence of the no-infill surrogates. Top: r2-score for $C_L$ (left) and $C_D$ (right); the dashed line marks the 0.99 target. Bottom: normalized RMSE for $C_L$ (left) and $C_D$ (right); the dashed line marks the 3\% target. Kriging, PODI and GNN are trained on DoE dataset with size ranging from 10-80 samples.}
        \label{fig:ClCdMetrics_vs_nSamples_AllGPBased}
    \end{figure}

    First, no-infill surrogates are built from pure DoE data in order to measure the raw predictive capability of each model. Three models are considered: a GP regressor for scalar quantities, a PODI model for field variables, and a GNN that operates directly on the surface-mesh graph. Separate data sets of 10–80 DoE  points are used for training, while 20 samples each are reserved for validation and testing. \autoref{fig:ClCdMetrics_vs_nSamples_AllGPBased} shows the prediction quality of the three surrogates as the DoE samples are gradually increased from 10 to 80 samples. A performance target of r2-score $\ge0.99$ and a error (normalized RMSE) below 3\% was set for scalar outputs (predicted or integrated from field predictions).
    The GNN reaches the r2-score threshold for both $C_L$ and $C_D$ with as few as 50 samples, whereas PODI requires roughly 60–70 samples and the GP (Kriging) about 70–80 samples. Similarly, GNN reaches the target error in 50-60 samples, followed by PODI with about 70 samples and GP requiring all 80 samples. While the GNN meets both accuracy goals with about 50 DoE samples, it lacks a rigorous built-in uncertainty estimate. Conversely, the GP provides an analytic variance but needs far more data to reach accuracy thresholds. These findings motivate an adaptive sampling (infill) stage after an initial DoE - (i) to accelerate GP convergence by targeting regions of highest predictive variance, and (ii) to supply GNN with strategically placed snapshots that refine the local flow physics without incurring the cost of a brute-force DoE expansion.

    \subsection{Surrogate Models with Adaptive Sampling}
        \begin{figure}[t]
            \centering
            \includegraphics[trim={0 0 0 0}, scale=0.55, clip]{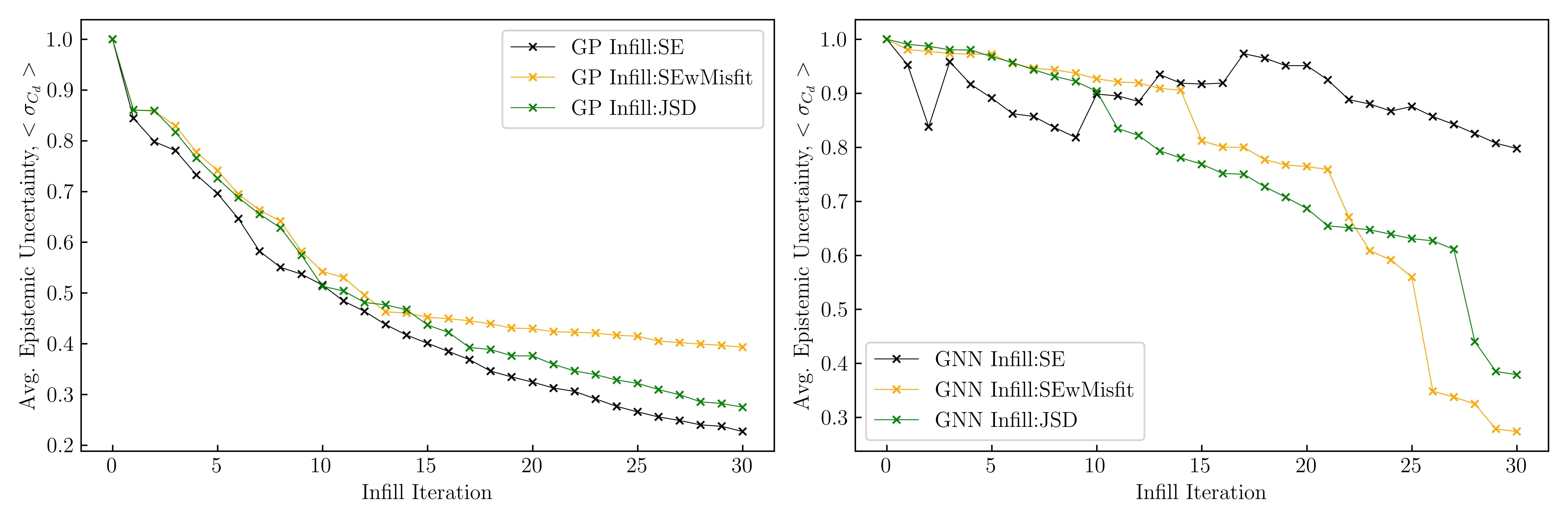}
            \caption{Reduction of scalar epistemic uncertainty (mean predictive variance $\langle\sigma_{C_D}\rangle$) under different infill approaches. Left: GP variance for Infill:SE, Infill:SEwMisfit and Infill:JSD. Right: GNN variance for Infill:SE and the corresponding GNN variance for the two coupled approaches. Values are averaged over 10,000 MC points in input space. Note that GP Infill:SE and GNN Infill:SE are independent approaches, and use GP and GNN surrogate errors, respectively.}
            \label{fig:AvgEpUnc_AllGPBasedInfill}
        \end{figure}
        \begin{figure}[!htbp]
            \centering
            \includegraphics[trim={0 0 0 0}, scale=0.55, clip]{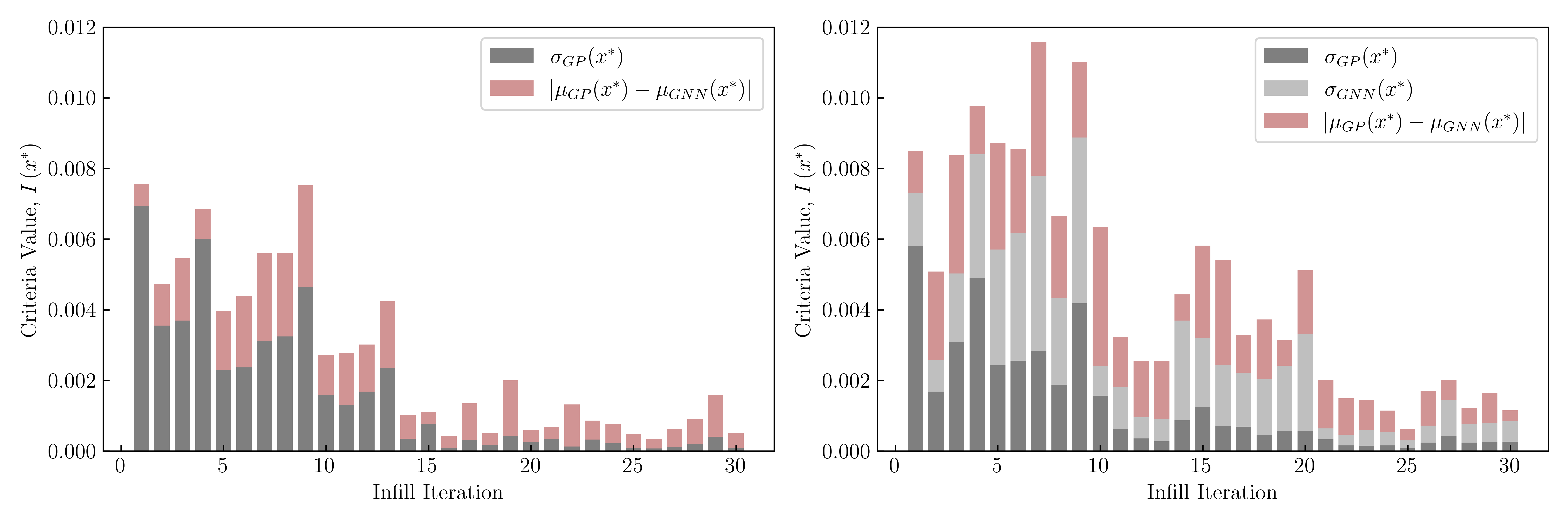}
            \caption{Evolution of the criteria value over infill iterations for different infill strategies with iteration 0 marking the end of the DoE stage. The criteria value is the term $I(x^*)$ in the expression of the general acquisition function  $\bm{\xi}^* = \argmin_{\bm{\xi}} \{- \, \text{PDF}_{\xi} \, I(\bm{\xi}^*)\}$. Stacked bars show contributions from different terms in the criteria value for SEwMisfit (left) and JSD (right) approaches. Such breakdown doesn’t directly relate to JSD criteria, but shown here for the sake of comparison with SEwMisfit.}
            \label{fig:InfillValue_AllGPBasedInfill}
        \end{figure}

        Next we employ adaptive sampling strategy. A common initial DoE set of 30 points is augmented by 30 infill samples that are acquired sequentially. Thus every adaptive surrogate is trained iteratively and monitored with the same validation and test sets as earlier. Four variants are updated after each infill step: a stand-alone scalar GP, a stand-alone GNN for the field outputs, and two coupled GP–GNN models (misfit and JSD). Here, we only rely on the GNN model for field predictions given its superior performs compared to PODI in the no infill stetting. However, the methodologies (at least SEwMisfit) can be applied to PODI models in a straightforward fashion. For uncertainty aware sampling w.r.t. the field model we would suggest to use GP model for latent space regression to also obtain uncertainty estimates for the PODI model.
        
        \autoref{fig:AvgEpUnc_AllGPBasedInfill} shows the scalar epistemic variance $<\sigma_{C_D}>$ for both GP and GNN, averaged over 10,000 MC points in the input space, as successive infill points are added.  
        For the GP surrogate, Infill:SE reduces epistemic uncertainty fastest to about 25\% of the initial spread, followed by JSD coupled infill strategy which reaches about 30\% of the initial level. The SEwMisfit infill approach reduces the GP surrogate average uncertainty to about 40\% of the initial.
        When the GNN is updated only using the MC-Dropout based estimate of uncertainty, its own variance remains high and erratic. However, the coupled criteria that combine GP variance with either GP–GNN misfit (SEwMisfit) or their Jensen-Shannon divergence reduce GNN's average uncertainty by roughly one order of magnitude. It is worth noting that beyond the $14^{th}$ iteration the uncertainty in GNN drastically reduces indicating the dominance of the misfit term in the acquisition function. The average uncertainty in GNN using JSD based infill is slightly higher compared to the SEwMisfit based approach. 

        \autoref{fig:InfillValue_AllGPBasedInfill} shows the progression of the criteria value of the acquisition function across infill iterations for the two coupled infill strategies. During the first 13 infill iterations the SEwMisfit criterion drops sharply with GP predictive variance dominating the criteria value. Over the remaining iterations the criteria value remain significantly smaller and is mostly dominated by the GP-GNN misfit.
        For the JSD based approach we stack the predictive variance from both surrogates along with the misfit. The criteria values drop more gradually as compared to SEwMisfit. The GNN predictive variance mostly dominates the value until 20 iterations, indicating high epistemic uncertainty in the neural network. 
        Since the JSD rewards overlap between the two predictive distributions rather than absolute error, it continues to propose samples that steadily reconcile GP and GNN even after the largest disagreements have been resolved. By iteration 30 SEwMisfit results in a lower overall criteria value while JSD maintains a slightly higher value. Both approaches suggest greater robustness and allow for the surrogate pair to gradually converge as compared to the SE based approach using either GP and GNN alone.

        \begin{figure}[t]
            \centering
            \includegraphics[trim={0 0 0 0}, scale=0.6, clip]{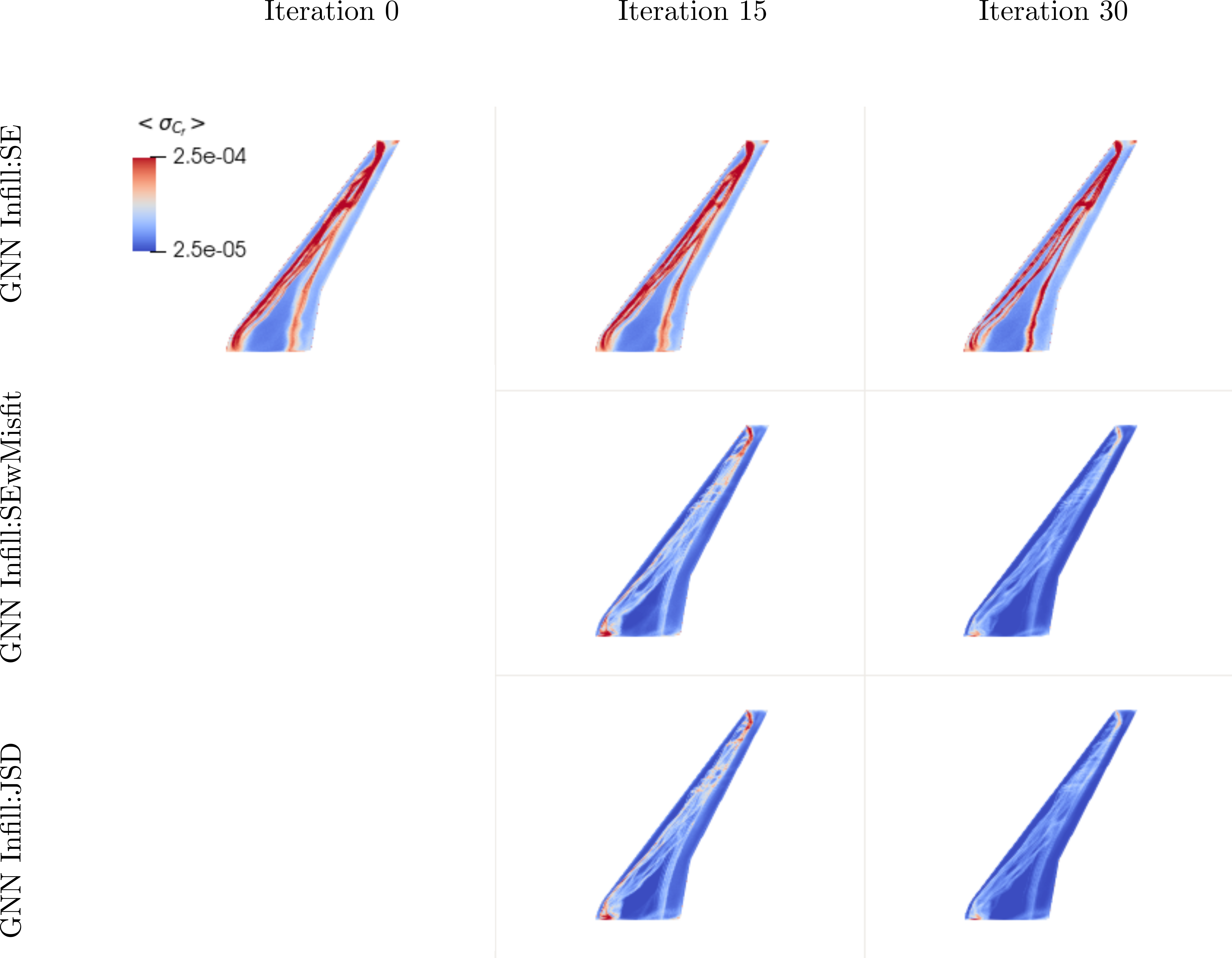}
            \caption{Spatial distribution of the skin-friction epistemic uncertainty $\sigma_{C_f}(x,y)$ on the CRM wing for the different infill strategies averaged over 100 input samples. Columns correspond to the initial DoE with 30 samples (i.e. infill iteration 0), infill iteration 15, and iteration 30. Rows correspond to the approaches Infill:SE, Infill:SEwMisfit and Infill:JSD.}
            \label{fig:AvgEpUncEstn_Cf}
        \end{figure}
         The uncertainty in the surface friction coefficient field is presented in \autoref{fig:AvgEpUncEstn_Cf}. At the beginning of the infill stage (iteration 0) large uncertainty around the shock and the highly-accelerated leading-edge region is observed. This can be attributed to the low number of DoE samples used for the initial GNN surrogate resulting in high variance predictions. By iteration 15, the coupled infill strategies have significantly reduced the uncertainty as compared to the SE based infill which leaves them virtually unchanged. After 30 infill points SEwMisfit and JSD move the uncertainty over the entire wing to a low variance regime, whereas the SE based approach still struggles to even mildly diminish the uncertainty. The coupled approach therefore ensures that remaining model epistemic uncertainty is sufficiently small to not contaminate the propagation of other sources of uncertainty. 

        \begin{figure}[t]
            \centering
            \includegraphics[trim={0 0 0 0}, scale=0.55, clip]{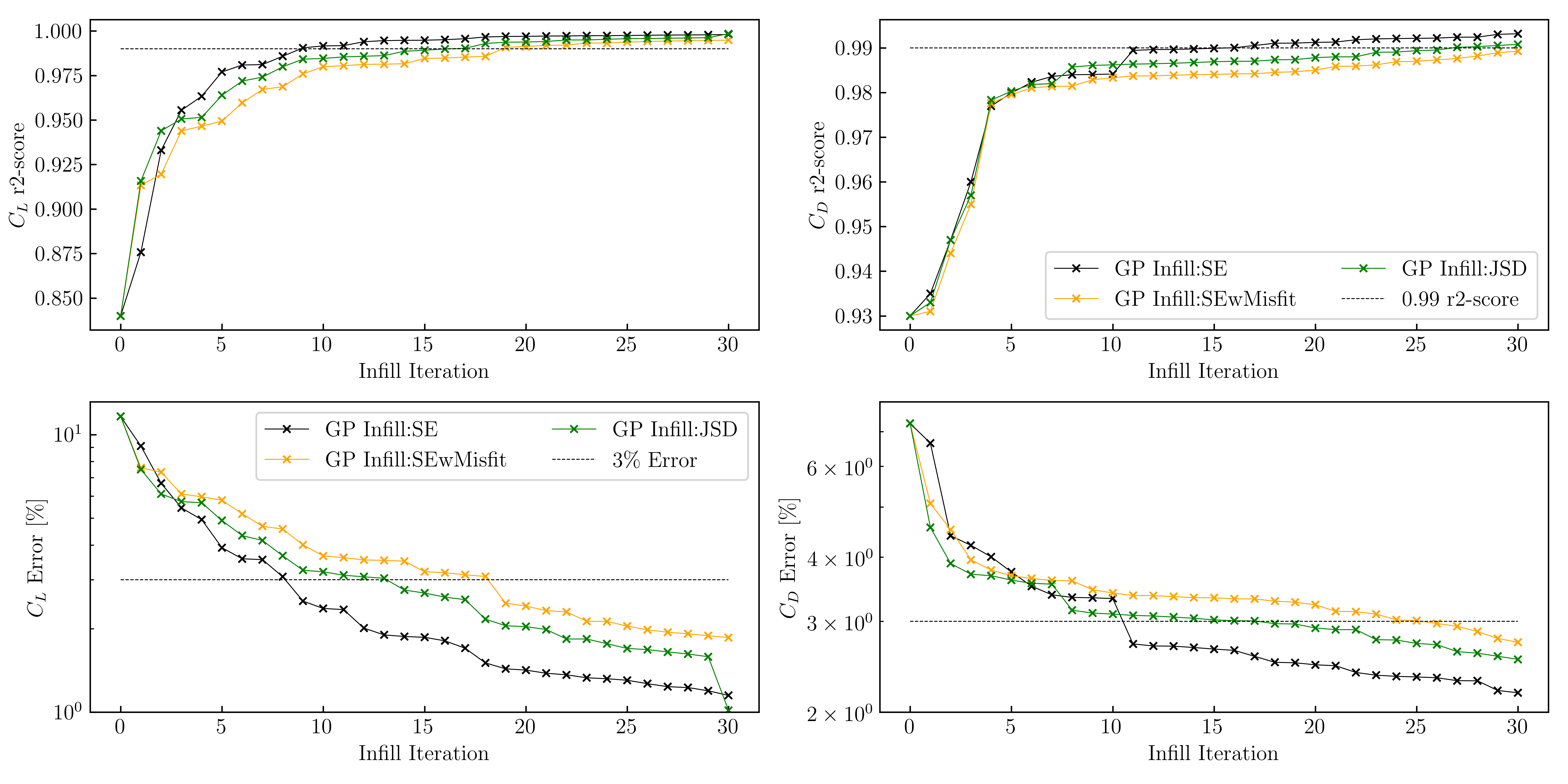}
            \caption{Evolution of GP accuracy under three acquisition rules. Upper row: coefficient of determination (r2-score) for lift (left) and drag (right); the dashed line marks the 0.99 target. Lower row: relative RMSE normalized by the data range for lift (left) and drag (right); the 3\% specification is indicated by the dashed line. Curves start at the 30 sample DoE (iteration 0).}
            \label{fig:ClCdMetrics_vs_nSamples_AllGPBasedInfill}
        \end{figure}
        
        \autoref{fig:ClCdMetrics_vs_nSamples_AllGPBasedInfill} summarizes how the GP surrogate predictions for the scalar outputs on the test dataset improves once adaptive sampling is activated. GP with SE-based infill reaches the target r2-score and error within 10 iterations. Whereas, GP with SEwMisfit and JSD infill takes about 30 and 23 iterations, respectively. Note that, in comparison to no-infill surrogates (\autoref{fig:ClCdMetrics_vs_nSamples_AllGPBased}), the adaptive sampling strategies tends to converge faster towards the target accuracy levels. 

        \begin{figure}[t]
            \centering
            \includegraphics[trim={0 0 0 0}, scale=0.55, clip]{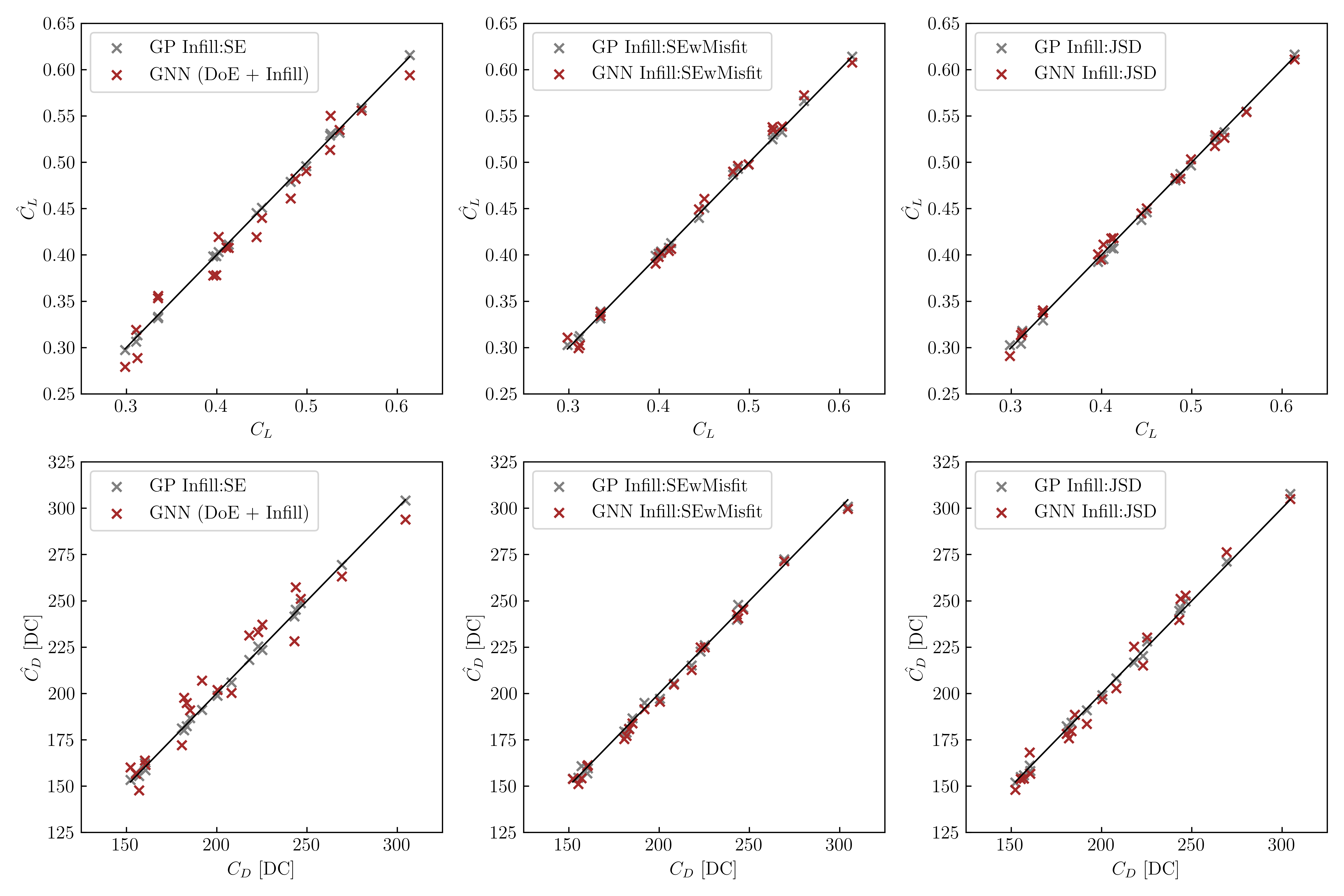}
            \caption{Parity plots for lift (top row) and drag counts (bottom row) after 30 adaptive sampling iterations. Columns correspond to the three acquisition strategies namely, variance–only (Infill:SE), coupled variance and misfit (Infill:SEwMisfit) and JSD based coupling (Infill:JSD).}
            \label{fig:ClCdRegPlot_AllGPBasedInfill}
        \end{figure}
        The predicted ($\hat{C}_L, \hat{C}_D$) versus true ($C_L, C_D$) scalar coefficients for the test dataset after 30 infill iterations are presented in \autoref{fig:ClCdRegPlot_AllGPBasedInfill}. The GNN (DoE + Infill) is trained on DoE samples plus SE based infill samples used to enrich GP. The GP scalar predictions using all three infill strategies are significantly closer to the true scalar values. Compared to GNN (DoE + Infill), the adaptive sampling based GNN scalar predictions (integrated using the field predictions) are much more accurate and closer to the GP predictions. This can be attributed to the reduction of the misfit term between GP and GNN during the infill iterations of the coupled strategies.    

        \begin{figure}[!t]
            \centering
            \includegraphics[trim={0 0 0 0}, scale=0.475, clip]{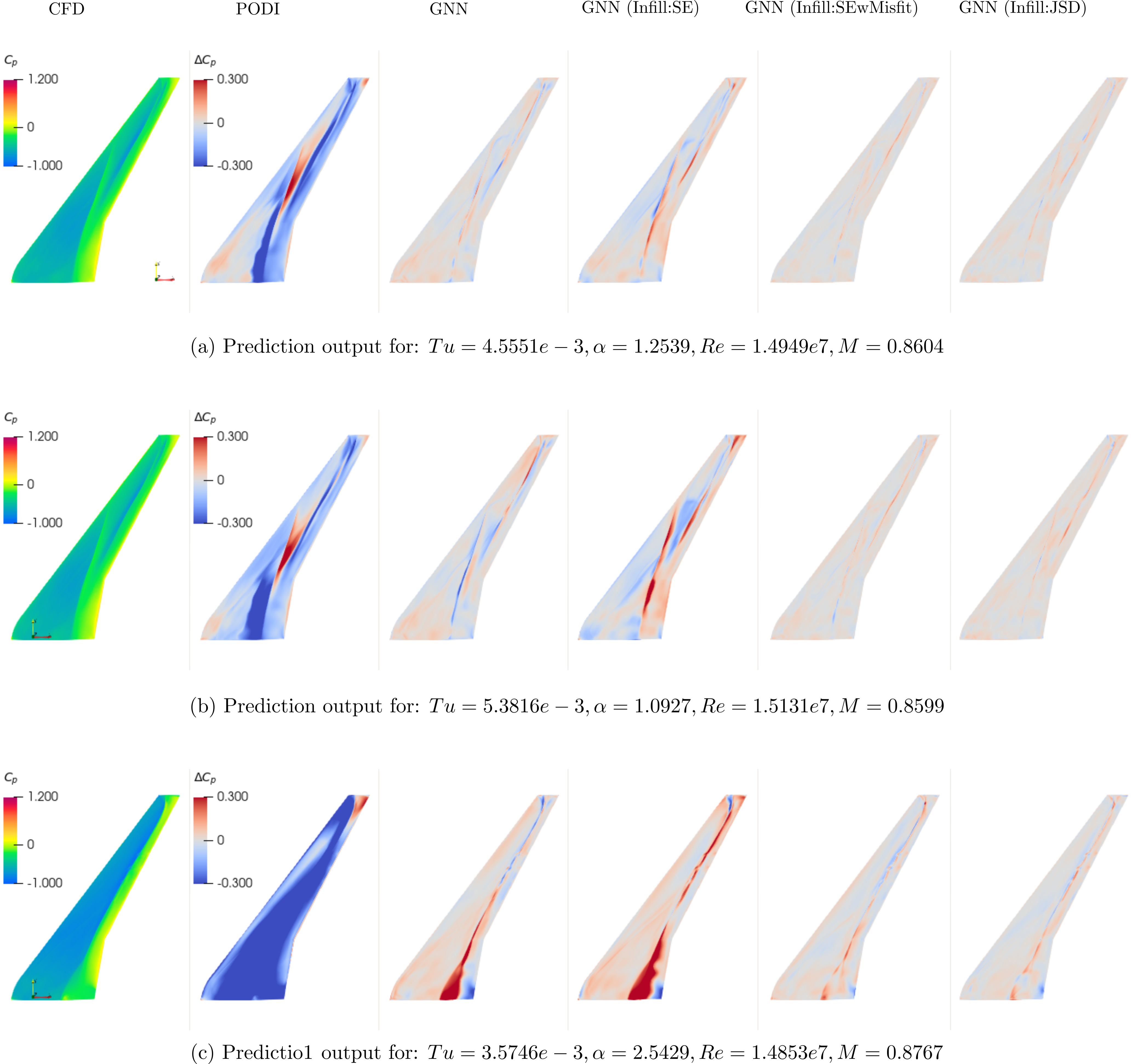}
            \caption{Surface pressure coefficient field predictions using different field surrogate models at three different test points. Points (a) and (b) belong to high probability region of the input space, while point (c) lies closer to the boundary of the joint distribution. The difference between surrogate and CFD output is plotted to clearly spot the prediction inaccuracies. Note that the model without infill are trained using 60 DoE samples.}
            \label{fig:all_predictions}
        \end{figure}
        
        In \autoref{fig:all_predictions}, we compare for three test points, the field predictions with the true field outputs for all the field surrogate models discussed previously. The model without infill are trained using 60 DoE samples. The difference between the predicted and the true fields is used to identify the regions of inaccuracies.
        The PODI fails to accurately capture the surface pressure coefficients. It shows large predictions errors close to and beyond the transition front, and near the shock locations as well. The no-infill GNN shows significant improvement over the PODI model but still fails to capture the transition front and shock locations accurately. GNN (Infill:SE) model performs better than PODI but still struggles to reach the accuracy of no-infill GNN. The coupled sampling based GNN models (Infill:SEwMisfit and Infill:JSD) have the highest accuracy of all the field surrogates. They have very low error along the transition front and are able to precisely capture the shock locations. Note that these are the only surrogate models which have a reasonable accuracy even for the test samples close to the edge of design space (\autoref{fig:all_predictions} (c)).     
    
        \begin{table}[t]
            \centering
            \footnotesize
            \begin{tabular}{lcccc}
            \toprule
            Method & $C_p$ RMSE & $C_f$ [x, y, z] RMSE in $e^{-4}$ & $C_p$ r2-score & $C_f$ [x, y, z] r2-score \\
            \midrule
            PODI & 0.1772 & 7.3816, 2.6253, 2.8166 & 0.7740 & 0.5073, 0.6691, 0.8892 \\
            \makecell[l]{GNN\\} & \textbf{0.0345} & 3.0214, 0.9096, 0.7151 & \textbf{0.9914} & \textbf{0.9176}, 0.9613, 0.9920 \\
            \makecell[l]{GNN (Infill:SE)} & 0.0627 & 5.2325, 1.5350, 1.1119 & 0.9717 & 0.7530, 0.8898, 0.9828 \\
            \makecell[l]{GNN (Infill:SEwMisfit)} & \textbf{0.0225} & 1.9923, 0.6403, 0.5204 & \textbf{0.9963} & \textbf{0.9641}, 0.9808, 0.9962 \\
            \makecell[l]{GNN (Infill:JSD)} & \textbf{0.0203} & 1.7977, 0.7124, 0.5535 & \textbf{0.9970} & \textbf{0.9708}, 0.9762, 0.9957 \\
            \bottomrule
            \end{tabular}
            \caption{Comparison of accuracy criterion for different active learning methods.}
        \label{tab:cost}
        \end{table}

        Error metrics for all models with respect to RMSE and r2-score are given in \ref{tab:cost} highlighting that the adaptive sampling strategies are outperforming the non-adaptive techniques with respect to all metrics consistently. Comparing the adaptive sampling criterions to each other it can be stated that the SEwMisfit as well as the JSD scheme yield very similar results while the SE approach is performing worse. Given that the latter does not account for both models simultaneously this does not come as a surprise.

\section{Uncertainty Propagation -- An engineering application}
    \begin{figure}[t]
        \centering
        \includegraphics[trim={0 0 0 0}, scale=0.7, clip]{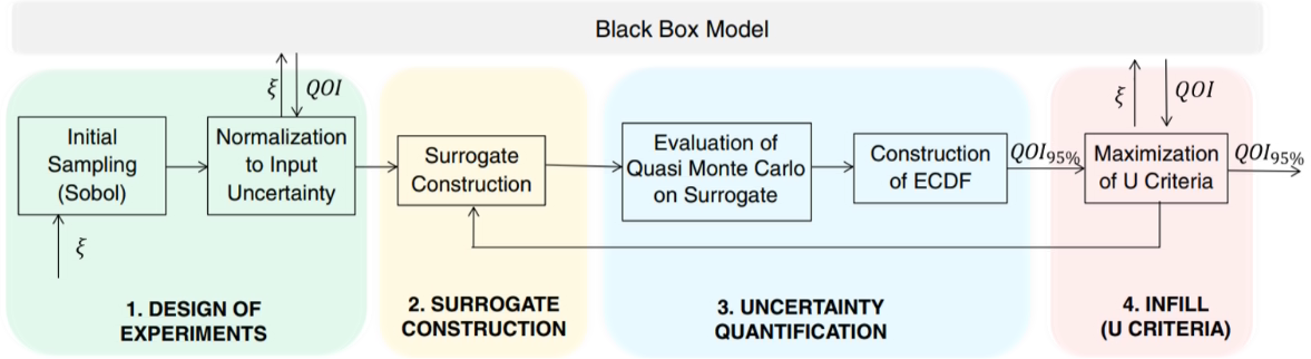}
        \caption{Steps involved in surrogate based uncertainty propagation approach.}
        \label{fig:SBUQ}
    \end{figure}

    We employ the Surrogate-Based Uncertainty Quantification (SBUQ) framework introduced by Sabater et al.\cite{Sabater2020} and available in SMARTy \cite{Bekemeyer2022}, as illustrated in \autoref{fig:SBUQ}. In this framework, a surrogate model maps the uncertain input variables $\boldsymbol{\xi}$ to a quantity of interest (QoI), enabling efficient propagation of input uncertainties. Rather than performing computationally intensive direct Monte Carlo simulations on the high-fidelity model, the framework uses a surrogate model to estimate statistical properties of the QoI. This strategy significantly reduces the computational burden by leveraging a fast-to-evaluate approximation of the true model response. To compute QoI statistics, the Monte Carlo method is employed not on the expensive black-box function, but on the surrogate model, which approximates its behavior. The surrogate model is trained on a set of initial Design of Experiments (DoE) samples in the stochastic input space. The predictive mean and an associated uncertainty (variance) estimate of the surrogate model, makes it well-suited for guiding adaptive sampling strategies. To enhance surrogate accuracy, we adopt an active learning (infill) strategy that sequentially refines the surrogate model. In the baseline version, the infill criterion focuses on improving global prediction of the QoI. Once the surrogate is refined, QoI statistics are computed using a large number of samples generated via Quasi-Monte Carlo (QMC) methods. Compared to conventional Monte Carlo, QMC offers improved convergence rates and better integration accuracy in high-dimensional settings, while maintaining ease of implementation.

      \begin{figure}[tp]
        \centering
        \includegraphics[trim={0 0 0 0}, scale=0.55, clip]{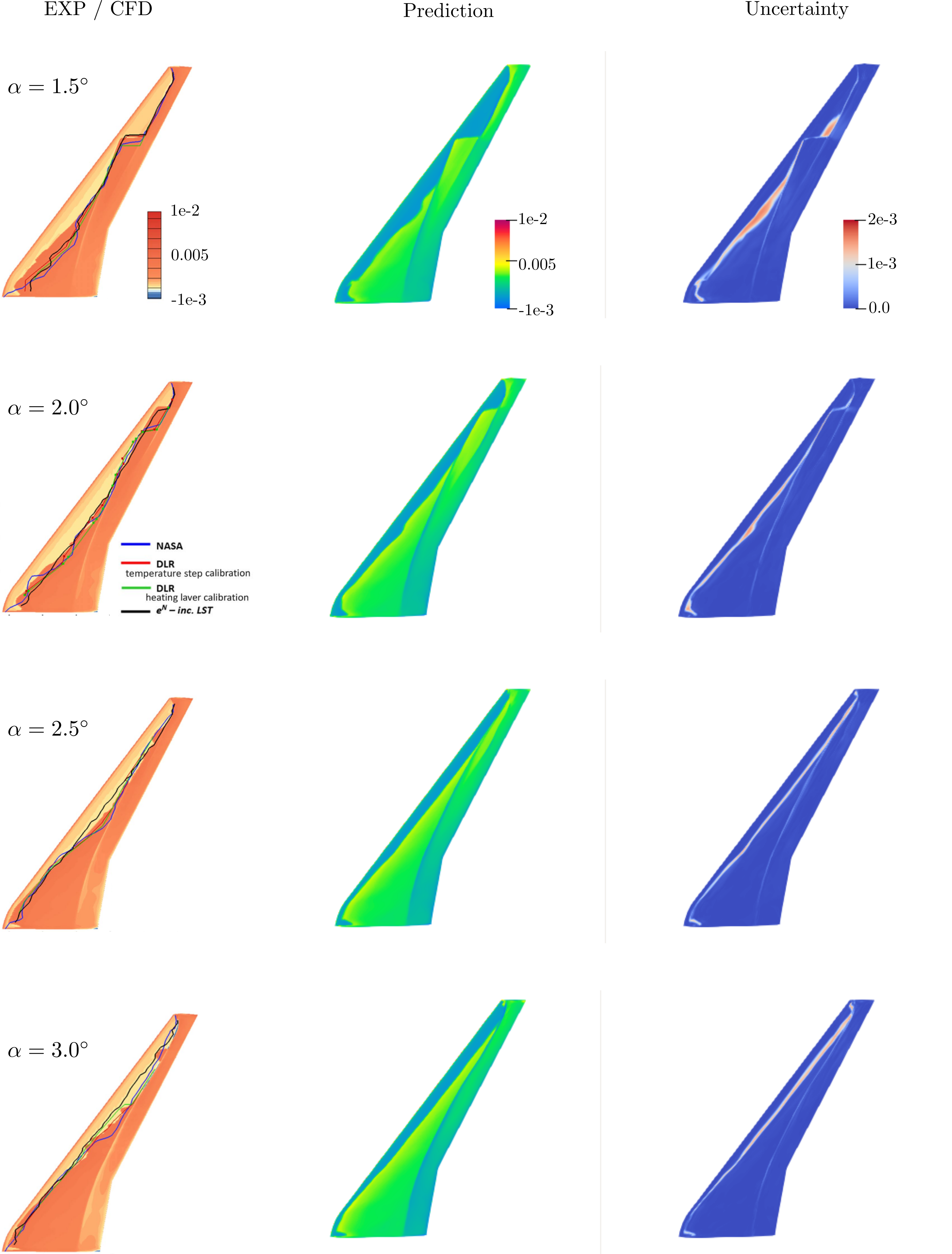}
        \caption{Surface skin friction coefficient at four different angles of attack. The CFD simulations with experimental data (first column) is adapted from \cite{Krumbein2022}. The nominal prediction and uncertainty reflected due to uncertain turbulent freestream  intensity is based on the SBUQ approach using the field surrogate with GP-GNN Infill:SEwMisfit as infill criterion.}
        \label{fig:SBUQ_Cf}
    \end{figure}
    \begin{figure}[tbp]
        \centering
        \includegraphics[trim={0 0 0 0}, scale=0.6, clip]{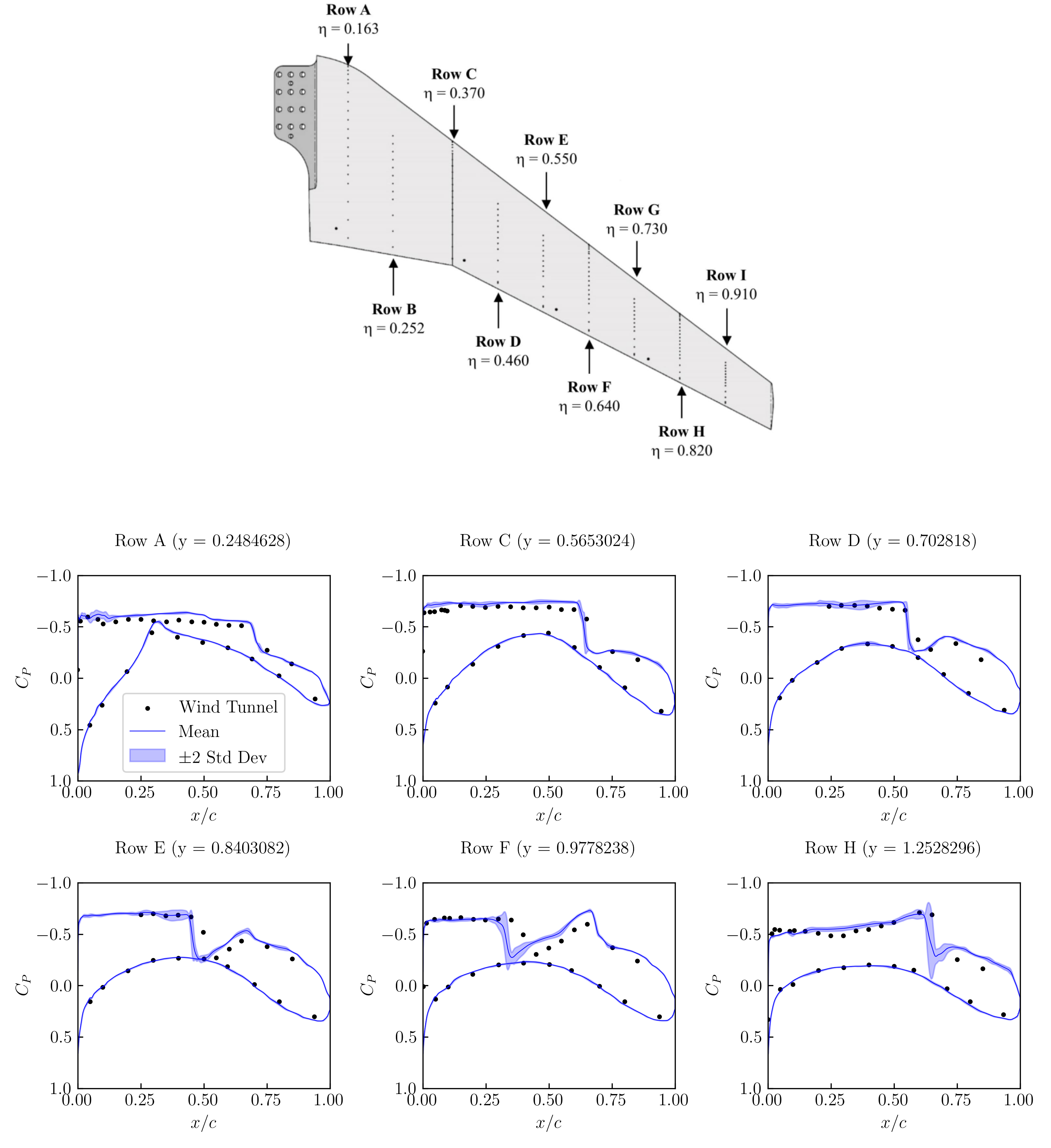}
        \caption{Surface pressure coefficient over different stations along the wingspan at an angle of attack of $1.5^{\circ}$. The uncertainty propagation results are indicated using the $\pm2$ standard deviation bounds around the surrogate mean prediction. The mean prediction and uncertainty propagation is based on the SBUQ approach using the field surrogate with GP-GNN Infill:SEwMisfit as infill criterion.}
        \label{fig:SBUQ_Cp}
    \end{figure}
    
    In contrast to \cite{Sabater2020} herein we substitute the scalar-valued based infill criterion with the previously discussed SEwMisfit criterion. This allows us to simultaneously propagate input uncertainties to the drag coefficient as well as the surface skin friction and pressure coefficients. While the drag coefficient provides insight into the aerodynamic performance, the skin friction and pressure coefficients enable us to investigate the influence of input uncertainties on transition locations and local aerodynamic behavior leading to a further improvement of our aerodynamic understanding. Similar to before we compute 30 DoE samples using a Halton sequence and add 30 infill samples based on the SEwMisfit criterion. Sampling is enhanced by accounting for the input PDFs to concentrate samples in locations of high likelihood. This results in the same model discussed in Sec.~\ref{sec:Results} labeled SEwMisfit but in addition we can investigate the influence of input uncertainties on distributed quantities by drawing a lot of samples from the GNN model using a quasi Monte Carlo evaluation. Note that, very similar results have been obtained when using the JSD criterion for infill.

    In \autoref{fig:SBUQ_Cf} the influence of the uncertain turbulent freestream  intensity on the surface skin friction coefficient is shown for four different angles of attack. The left column shows a combination of experimental and numerical results and is adapted from \cite{Krumbein2022}. Note the, changes in transition locations derived based on different criteria/approaches. The section column displays the mean surface skin friction coefficient while its variance is displayed in the third column. A close correlation between mean prediction and the combined EXP/CFD is clearly visible. Interestingly the variation in turbulent freestream intensity results in a variance around the transition location that resembles the differences in transition location derived based on changing criteria. Transition for angles of attack above 2.5 degrees are primarily caused from the occurring shock on the upper surface. Since the influence of the turbulent freestream intensity is minimal, also the predicted variance shrinks compared to smaller angles of attack.

    In addition to the skin friction coefficient also the surface pressure coefficient under uncertainties can be analyzed and compared to experimental results. \autoref{fig:SBUQ_Cp} shows the $c_p$ distribution at six different stations along the wingspan for an angle of attack of $1.5^{\circ}$  with experimental results as dots and the mean prediction as solid lines. Variances are indicated by shaded areas. Overall reasonable agreement between the model predictions and experimental results is observed. For the most inboard section Row A the effect of variations in turbulent intensity is occurring towards the leading edge at around $x/c = 0.1$ which relates to variations seen in surface skin friction coefficient in \autoref{fig:SBUQ_Cf}. For Row C and D there is barely any influence of the turbulent freestream intensity visible whereas for more outboard sections variance are primarily present in the proximity of the shock. Interestingly the uncertainty in turbulent freestream intensity seems to cause a variation in shock intensity but seems to have a very limited influence on the shock location.

\section{Conclusion}\label{sec:Conclusion}
    In this paper we present a goal-driven adaptive sampling strategy for models predicting fields such as the surface pressure distribution of an aircraft. The proposed strategy is agnostic to the field prediction model at hand and hence suitable for arbitrary deep-learning models. Effectively it consists of two components that a combined to form a composite infill criterion. First, a Gaussian Process which inherent error estimate is used to determine undersampled areas of the domain and, second, a misfit term between the integrated field model and Gaussian process prediction ensuring consistency between both models. The misfit term can either be computed in a deterministic fashion by comparing mean prediction or in a stochastic fashion by leveraging the Jensen-Shannon Divergence term.

    The performance of the newly proposed adaptive sampling strategy is investigated for the NASA Common Research Model comparing the model prediction accuracy for different infill strategies and also a baseline quasi-monte Carlo, model-agnostic space-filling sampling strategy. The different strategies are infill solely-based on the Gaussian Process error and infill based on the composite formulations, namely surrogate error with miswfit (SEwMisfit) as well as a stochastic formulation based on the Jensen-Shannon Divergence term (JSD). Graph neural networks are used a field-prediction models while also some initial results are shown for proper orthogonal decomposition with latent-space interpolation. Both composite infill criteria, SEwMisfit and JSD, consistently outperform the GP-only based criterion with respect to field as well as scalar value predictions. It has been observed that first the GP error portion of the criterion is active ensuring that the entire domain is adequately covered while in the later stage of the infill process the misfit term dominates and effectively refines critical flow regions, e.g. high angles of attack and Mach number, where a mismatch between local and global predictions is likely. Finally the SEwMisfit criterion is employed for forward uncertainty propagation analysis investigating the influence of uncertainties in turbulent intensity on the surface pressure and skin friction coefficient of the natural laminar flow version of the NASA CRM. Results coincide with experimental findings with respect to pressure distributions and derived laminar to turbulent transition location.

    The proposed methodology could further be leveraged in other engineering fields and/or combined with different deep-learning model. Given that the ability to combine the distributed field to an ideally engineering meaningful value is the only prerequisite, the proposed strategy is rather flexible and can be easily adopted to other fields, e.g. adaptive sampling for aerodynamic performance and load databases. Moreover, it would be interesting to investigate more rigorous uncertainty quantification methods for the employed graph neural networks since it has been observed that the drop-out based method employed herein sometime resulted in rather erratic behavior throughout the infill process.

\begin{appendix}
\section{Estimating Epistemic Uncertainty in GNN}\label{app:GNNEpUnc}
    Epistemic uncertainty in a model can be described as the uncertainty in the model's predictions due to the lack of knowledge about the underlying model parameters. In neural networks, this can be approximated by using Monte Carlo (MC) dropout during inference, where the dropout is kept active even during the prediction phase. The procedure for approximating epistemic uncertainty adopted in this study is as follows:

    \begin{itemize}
        \item The GNN is used in training mode to generate different outputs for the same input \( x \).
        \item For each input, \( 10,000 \) MC dropout samples of the field output are generated, which are in turn integrated to obtain the desired scalars (aerodynamic coefficients). The mean \( \mu_{GNN}(x) \) and variance \( \sigma_{GNN}(x) \) of the scalar outputs are then computed.
    \end{itemize}

    The use of MC dropout for estimating epistemic uncertainty in the strategy discussed above can slow down computations during optimization in surrogate models. Specifically, using MC dropout during each infill step significantly increases the computational burden, as estimates of \( \mu_{GNN}(x_i) \) and \( \sigma_{GNN}(x_i) \) must be computed at several input points \( x_i \). 
    To mitigate this issue, we pre-compute two surrogate models before each infill step for the mean and and standard deviation using 1000 input points, and use them during optimization. This reduces computational time by eliminating the need of performing MC dropout at thousands of probe locations during optimization.
    
\end{appendix}

\section*{Acknowledgments}\label{sec:Acknowledgments}
The funding of parts of these investigations from the Deutsche Forschungsgemeinschaft (DFG, German Research Foundation) under Germany’s Excellence Strategy—EXC 2163/1-Sustainable and Energy Efficient Aviation—Project-ID 390881007 is gratefully acknowledged. 

The authors gratefully acknowledge the scientific support and HPC resources provided by the German Aerospace Center (DLR). The HPC system CARA is partially funded by 'Saxon State Ministry for Economic Affairs, Labor and Transport' and 'Federal Ministry for Economic Affairs and Climate Action'. The HPC system CARO is partially funded by "Ministry of Science and Culture of Lower Saxony" and "Federal Ministry for Economic Affairs and Climate Action". 

\bibliographystyle{elsarticle-num-names} 
\bibliography{AST2025}





\end{document}